\documentclass[runningheads]{llncs}
\usepackage{graphicx}
\usepackage{comment}
\usepackage{amsmath,amssymb} 
\usepackage{color}
\usepackage{url}
\usepackage{subfigure}

\usepackage[width=122mm,left=12mm,paperwidth=146mm,height=193mm,top=12mm,paperheight=217mm]{geometry}

\begin{document}

\pagestyle{headings}
\mainmatter
\def\ECCVSubNumber{100}  

\title{Face Anti-Spoofing by Learning Polarization Cues in a Real-World Scenario} 


\author{Yu Tian, Kunbo Zhang, Leyuan Wang, Zhenan Sun}
\institute{{ Institute of Automation, Chinese Academy of Sciences \\
	Center for Research on Intelligent Perception and Computing \\
    Tianjin Academy for Intelligent Recognition Technologies\\
    {\tt\small \{yu.tian,kunbo.zhang\}@ia.ac.cn,\{leyuan.wang\}@cripac.ia.ac.cn,\\
    \{znsun\}@nlpr.ia.ac.cn}}
}

\maketitle

\begin{abstract}
Face anti-spoofing is the key to preventing security breaches in biometric recognition applications. Existing software-based and hardware-based face liveness detection methods are effective in constrained environments or designated datasets only. Deep learning method using RGB and infrared images demands a large amount of training data for new attacks. In this paper, we present a face anti-spoofing method in a real-world scenario by automatic learning the physical characteristics in polarization images of a real face compared to a deceptive attack. 

\end{abstract}

\section{Introduction}

As one of the prevailing biometric authentication methods, face recognition technology has been widely used due to its convenience and accuracy. However, the challenges of various spoofing attacks jeopardize the security of personal information in such systems. Criminals can easily impersonate users to gain access to the system by using photos and videos with fake registered users information at a low cost. There is a strong demand for robust and accurate face anti-spoofing countermeasures to overcome the continuous improvement of attacks in complicated situations.

Researches on  faces anti-spoofing have been extensively carried out that many hardware-based and software-based methods are proposed\cite{hadid2014face}\cite{jourabloo2018face}. These methods are very reliable in detecting known spoofing attacks under controlled environment. When the detection environment changes or new attacking methods appear, the performance of such methods becomes awkward. It is known that spoofing attack is significantly different from human face skin in material, texture, surface roughness and other attributes, which will be apparent if the reflection and refraction light is captured by a polarized sensor. We believe that polarization imaging technology has the potential to overcome existing issues in face liveness detection applications.

\section{Related Work}
In this paper, we divide face liveness detection methods into hardware-based and software-based.

The software based anti-spoofing attack detection method has the characteristics of low cost and high accuracy, which has been developed rapidly in recent years. Early software methods require users to blink\cite{pan2007eyeblink}, turn their heads\cite{kollreider2005evaluating} or read according to instructions\cite{chetty2004liveness}, which can effectively respond to print attacks, but the user experience is poor, and it can not respond to video playback attacks. In order to solve these problems, the analysis method based on handcrafted feature is proposed in\cite{bharadwaj2014face}\cite{maatta2011face}\cite{boulkenafet2016face}. Although these methods can perform well in the specified dataset, they are not effective in practical applications. With the increasing number of public benchmark datasets\cite{zhang2018dataset}\cite{claeskens2014multivariate}\cite{yuan2008recent}\cite{zhang2018casia}, competitions for liveness detection have been held\cite{CVPR}, and more and more algorithms have been proposed\cite{george2019biometric}, the accuracy of detection is constantly refreshed. However, due to the small size of the existing datasets, and the types of spoofing attacks are endless, the generalization ability of such methods has not been effectively solved.

Compared with the software based method, the hardware based method uses a special sensor for image acquisition, which makes the difference between the genuine face and the spoofing attacks more prominent, so the detection effect is more stable. Due to the difference in reflectivity between real face and spoofing attack, multispectral\cite{zhang2011face}\cite{raghavendra2017face}, infrared\cite{sun2016context} and remote Photo Plethysmography (rPPG) methods\cite{hernandez2018time}\cite{liu20163d} are used in liveness detection, which have high accuracy, but the collection conditions are relatively strict, and the hardware system is relatively complex, so it is difficult to be widely used. In addition to reflecting information, depth based devices are also used for liveness detection, such as structured light\cite{connell2013fake} and TOF camera\cite{wu2019review}. These method can effectively deal with 2D plane attacks but not 3D attacks. \cite{sepas2017face}\cite{kim2014face} provides the method of liveness detection with light field camera, which can detect a variety of spoofing attacks, but the light field imaging equipment is expensive, and the detection results are seriously affected by the light.

\section{Proposed Method}

\begin{equation}
\centering
S = \left[ {\begin{array}{*{20}{c}}
{{S_0}}\\
{{S_1}}\\
{{S_2}}\\
{{S_3}}
\end{array}} \right] = \left[ {\begin{array}{*{20}{c}}
I\\
Q\\
U\\
V
\end{array}} \right] = \left[ {\begin{array}{*{20}{c}}
{{I_0} + {I_{90}}}\\
{{I_0} - {I_{90}}}\\
{{I_{135}} - {I_{45}}}\\
{{I_L} - {I_R}}
\end{array}} \right]
\end{equation}

Where, I is the sum of 0 degree polarized image (${I_0}$) and 90 degree polarized image (${I_{90}}$), Q represents the difference between ${I_0}$ and ${I_{90}}$, U is the difference between 45 degree polarized image (${I_{45}}$) and 135 degree polarized image (${I_{135}}$), V is the difference between left-handed circular polarized image(${I_L}$) and right-handed circular polarized image(${I_R}$). In the actual case, circular polarization and ellipsometry are rare, so the Stokes parameters is denoted as I, Q, U in the subsequent calculation process herein, and the V component is omitted.

After the Stokes parameter is obtained, the expression of DOLP image can be defined\cite{vedel2011full}, as shown in formula (2).
\begin {equation}
\centering
DOLP = \sqrt {\frac{{{Q^2} + {U^2}}}{I}}
\end{equation}

\subsubsection{Experimental Results}

\begin{figure}
\centering
\includegraphics[scale=0.4]{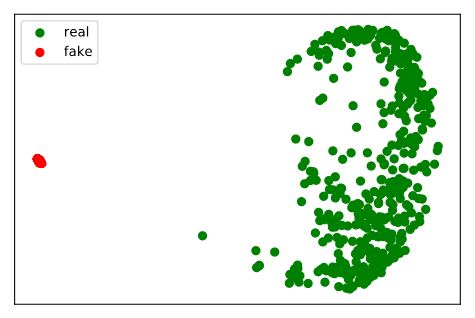}
\caption{The clustering results of genuine and fake face eigenvectors.}
\end{figure}

We verified the method in the Face-DOLP dataset and obtained very high accuracy results. In order to prove that this is due to the help provided by the polarization information rather than the introduction of the neural network, we also performed the same experiments on the grayscale and RGB datasets as shown in Table 3. We demonstrate the results of equal error rate (EER) and corresponding true positive rate (TPR) when the false positive rate (FPR) is 10e-3 and 10e-2.

\begin{figure}
\centering
\includegraphics[scale=0.5]{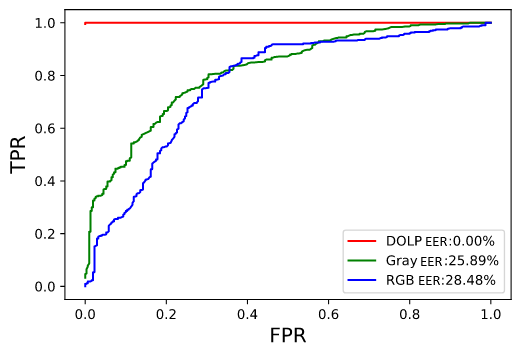}
\caption{Comparison of ROC curves for DOLP, grayscale and RGB images. It is obvious the advantages of using DOLP information for face anti-spoofing over traditional RGB and grayscale information with 0 EER.}
\end{figure}

\setlength{\tabcolsep}{4pt}
\begin{table}
\begin{center}
\caption{Evaluation results using DOLP, grayscale and RGB images based on MobileNet V2 network structure.}
\label{table:headings}
\begin{tabular}{llll}
\hline
Dataset & EER & TPR@FPR=10e-2 & TPR@FPR=10e-3\\
\hline
RGB & 28.48\% &0.2837 & 0.0186\\

GRAY & 25.88\% &0.4744 & 0.2674\\

\textbf{DOLP} & \textbf{0.0}\% &\textbf{1.0} & \textbf{1.0}\\
\hline
\end{tabular}
\end{center}
\end{table}
\setlength{\tabcolsep}{1.4pt}

In Figure 7, we show the receiver operating characteristic (ROC) cuvre. In the grayscale and RGB datasets, as the FPR increases, the TPR also starts to increase, but in the DOLP dataset, as the FPR increases, the TPR is always stable at 1.

In addition to verifying the superiority of PAAS in different datasets, we show the comparison of the experimental results of PAAS with existing polarization methods and handcrafted feature methods in Table 4, we compared with the three statistical methods of mean, standard deviation and kurtosis described in\cite{aziz2017face} and LBP features. In general, the detection results of PAAS in the common detection environment far exceed the manual feature extraction method and the existing polarization face anti-spoofing method.

Existing liveness detection methods are mainly aimed at research conducted under indoor visible light illumination, but face recognition systems at night or outdoors will still suffer spoofing attacks. As shown in Figure 8, we performed relevant experiments in these two unconventional environments, and PAAS can still perform stable detection.

\begin{figure}
\centering
\subfigure{
\begin{minipage}[t]{0.2\linewidth}
\includegraphics[width=0.9in]{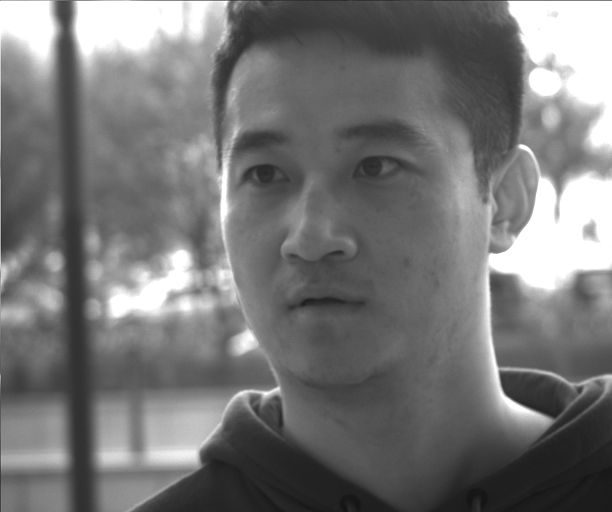}
\end{minipage}%
\begin{minipage}[t]{0.2\linewidth}
\includegraphics[width=0.9in]{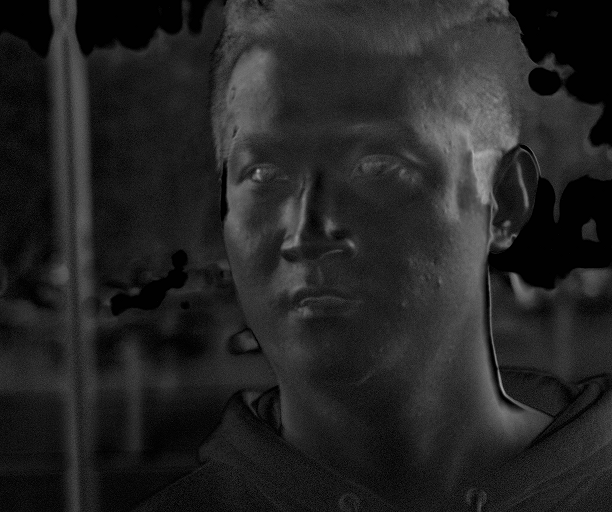}
\end{minipage}%
\begin{minipage}[t]{0.2\linewidth}
\includegraphics[width=0.9in]{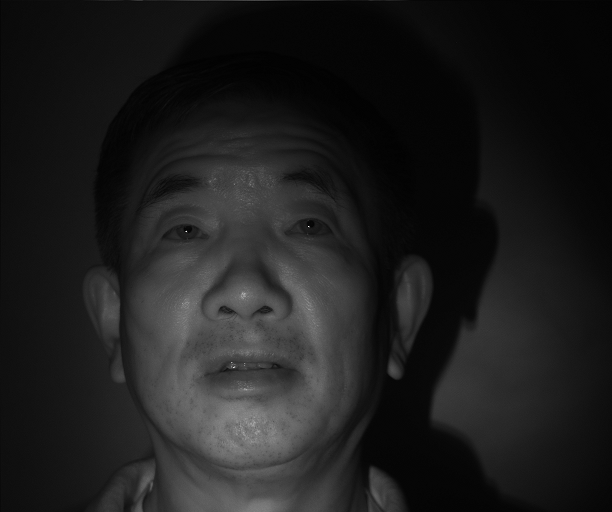}
\end{minipage}%
\begin{minipage}[t]{0.2\linewidth}
\includegraphics[width=0.9in]{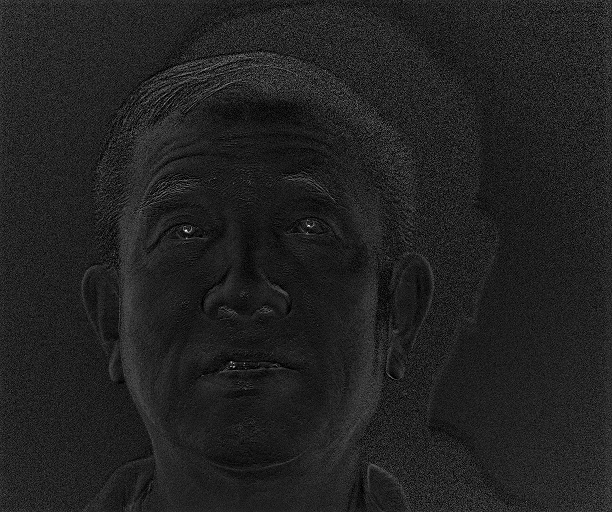}
\end{minipage}%
}%
\centering
\caption{Experimental results of PAAS method is investigated in outdoor and night scenarios. From left to right, the I0 image captured outdoor during daytime and its corresponding DOLP image, the I0 image taken at night without ambient light and its corresponding DOLP image.}
\end{figure}

PAAS not only has strong environmental adaptability, but also has strong generalization ability. We have done a lot of experiments on unknown subjects, all of which can get accurate detection results. Relevant experimental results are provided in supplementary materials in the form of video demos.

\setlength{\tabcolsep}{4pt}
\begin{table}
\begin{center}
\caption{Comparison of PAAS method based on deep learning with traditional handcrafted feature extraction methods including mean, standard deviation, Kurtosis and LBP.}
\label{table:headings}
\begin{tabular}{llll}
\hline
Method & EER & TPR@FPR=10e-2 & TPR@FPR=10e-3\\
\hline
Mean & 31.07\% &0.1814 & 0.0233\\

Standard deviation & 49.84\% &0.1535 & 0.0140\\

Kurtosis & 43.04\% &0.1628 & 0.0023\\

LBP & 32.04\% &0.5793 & 0.8803\\

\textbf{PAAS} & \textbf{0.0}\% &\textbf{1.0} & \textbf{1.0}\\
\hline
\end{tabular}
\end{center}
\end{table}
\setlength{\tabcolsep}{1.4pt}

\section{Conclusion}
In this paper, we propose a polarization face anti-spoofing (PAAS) method using a deep learning computational model to learn physical cues in polarization images.  In future research, we will further investigate the relationship between face physical characteristics and polarization information by interpreting the learning mechanism inside the convolutional neural network of the proposed PAAS method. It is also expected to challenge our method with vivid mask attacks using low cost polarization imaging hardware other than the sony sensor off the shelf.

\par\vfill\par

\clearpage
%
%
\bibliographystyle{splncs04}
\bibliography{mybib}
\end{document}